%% file: MainFile.tex
\documentclass[runningheads]{llncs}
\usepackage{geometry}
\usepackage{graphicx}
\usepackage{subcaption}
\usepackage{caption}
\usepackage{makecell} 
\usepackage{url}
\usepackage{gensymb}
\usepackage{float}

\usepackage{array}
\usepackage{tabularx,multirow}
\usepackage{mwe} 
\usepackage{subcaption}
\usepackage{float}
\usepackage{hyperref}
\makeatletter
\usepackage[symbol]{footmisc}
\usepackage{afterpage}
\usepackage{amsmath} 

\newcommand{\eg}[0]{\textit{e.g.},~}
\newcommand{\ie}[0]{\textit{i.e.},~}
\newcommand{\printfnsymbol}[1]{%
  \textsuperscript{\@fnsymbol{#1}}%
}
\makeatother
\begin{document}
%

\newcolumntype{x}[1]{>{\centering\let\newline\\\arraybackslash\hspace{0pt}}p{#1}}

\raggedbottom

\title{Addressing catastrophic forgetting for medical domain expansion}


\author{Anonymous}
\institute{Anonymous Organisation\\}

\author{Sharut Gupta\thanks{These authors contributed equally.}\inst{1,2} \and
Praveer Singh\printfnsymbol{1}\inst{1} \and
Ken Chang\printfnsymbol{1} \inst{1,3} \and
Liangqiong Qu \inst{4} \and
Mehak Aggarwal \inst{1} \and
Nishanth Arun \inst{1} \and
Ashwin Vaswani \inst{1} \and
Shruti Raghavan \inst{1} \and
Vibha Agarwal \inst{1,3} \and
Mishka Gidwani \inst{1} \and
Katharina Hoebel \inst{1,3} \and
Jay Patel \inst{1,3} \and
Charles Lu \inst{1} \and
Christopher P. Bridge \inst{1} \and
Daniel L. Rubin \inst{4} \and
Jayashree Kalpathy-Cramer \inst{1}}

\institute{Athinoula A. Martinos Center for Biomedical Imaging, Department of Radiology, Massachusetts General Hospital, Boston, MA, USA\\
\email{jkalpathy-cramer@mgh.harvard.edu}\\
\url{https://qtim-lab.github.io/}\and
Indian Institute of Technology Delhi, New Delhi, India \and
Massachusetts Institute of Technology, Cambridge, MA, USA\and
Department of Radiology and Biomedical Data Science, Stanford University, Palo Alto, CA, USA}

\maketitle              
 \begin{abstract}
Model brittleness is a key concern when deploying deep learning models in real-world medical settings. A model that has high performance at one institution may suffer a significant decline in performance when tested at other institutions.  While pooling datasets from multiple institutions and re-training may provide a straightforward solution, it is often infeasible and may compromise patient privacy. An alternative approach is to fine-tune the model on subsequent institutions after training on the original institution. Notably, this approach degrades model performance at the original institution, a phenomenon known as \textit{catastrophic forgetting}. In this paper, we develop an approach to address catastrophic forgetting based on elastic weight consolidation combined with modulation of batch normalization statistics under two scenarios: first, for expanding the domain from one imaging system's data to another imaging system's, and second, for expanding the domain from a large multi-institutional dataset to another single institution dataset. We show that our approach outperforms several other state-of-the-art approaches and provide theoretical justification for the efficacy of batch normalization modulation. The results of this study are generally applicable to the deployment of any clinical deep learning model which requires domain expansion.

\end{abstract}
\input{Introduction}
\input{Results}
\input{Discussion}
\input{Conclusion}
\input{Method}

\bibliographystyle{splncs04}
\bibliography{EWC}
\newpage
\input{Acknowledgments}
\newpage
\input{Supplement}

\end{document}

%% file: Introduction.tex
\section{Introduction}
Deep learning (DL) models have shown state-of-the-art performance for a wide variety of computer vision \cite{chen2020simple,tan2020efficientdet,chen2018encoder,DeepTMO}, biomedical signal processing \cite{jing2020development,ribeiro2020automatic}, and medical imaging tasks \cite{esteva2017dermatologist,li2020siamese}. Within the clinical context, there is a need to continually refine these models to achieve high performance on new datasets, such as those from different image acquisition systems or hospitals or institutions. Typically, DL models are extended to new datasets via fine-tuning the model weights; a neural network trained on the original dataset is used to initialize a new model, which is then trained on the target domain \cite{mohamed2017finetune}. However, this can result in \textit{catastrophic forgetting} of the previous dataset, a phenomenon in which models do not preserve previously learned knowledge and consequently result in a degradation of the performance on the original dataset \cite{goodfellow2013empirical}. This poses a major challenge for regulatory agencies, such as the Food and Drug Administration (FDA) in the United States, as DL models that have been fine-tuned after their approval may no longer satisfy the required performance criteria on the original test set. In this work, we explore techniques to mitigate catastrophic forgetting in the setting of fine-tuning on new medical datasets, also known as domain expansion.

Prior work in addressing domain expansion have focused on interventions that target specific layers of a network, \eg by either fine-tuning trainable parameters of the Batch Normalization (BN) layers \cite{karani2018lifelong} or modulating the BN statistics (bias/variance) \cite{li2016revisiting}; both of them attempting to rectify any differences in internal co-variate shift by aligning the distribution of the new dataset with the previous one, resulting in similar model performance across both datasets. However, these techniques necessarily limit the model's capacity to incorporate new knowledge since all non-BN layers are kept frozen throughout the fine-tuning process.

\begin{figure}[H]
\centering
    \subfloat[]{\includegraphics[width=\columnwidth]{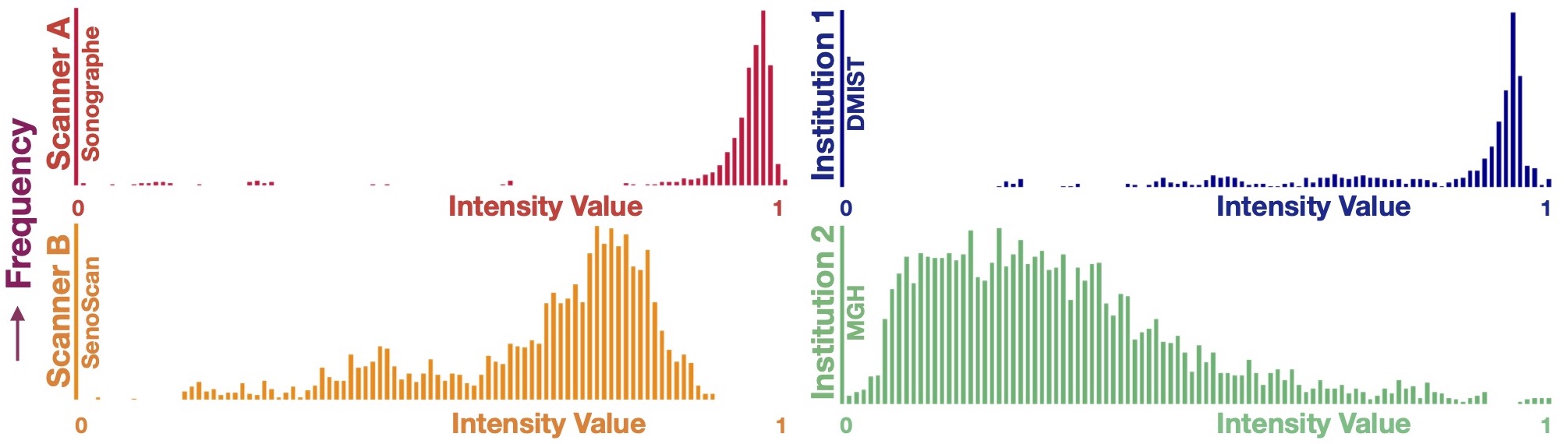}}
  \hfill
   \subfloat[\label{tab:subkey}]{%
       \begin{tabular}{ |c|c|c|x{5cm}|x{4cm}|c|x{1cm}| } 
 \hline
 Dataset &  Name & Composition & Acquisition System & Name &  Size & Total Size \\ \hline
 
 \multirow{10}{*}{\textbf{Institution 1}}  & \multirow{10}{*}{DMIST}   &   \textbf{Scanner A} &  General Electric Medical Systems &  SenoGraphe & 59411 & \multirow{8}{*}{103890}  \\ \cline{3-6}
    & & - & Fuji Medical & Computed Radiography System& - & \\ \cline{3-6}
    & & - & Hologic Mammography System &  Digital Mammography System & - & \\\cline{3-6}
    & & - &  Hologic  Mammography System &   Selenia Full Field System & - & \\ \cline{3-6}
    & & \textbf{Scanner B} &  Fischer Medical &   SenoScan & 32928 & \\ \hline
 \textbf{Institution 2} & MGH & - & Hologic mammography system & Lorad Selenia & 8603 & 8603\\ 
 \hline
\end{tabular}}

\caption{(a) Intensity histograms of 100 randomly selected images from each category showing heterogeneity across digital mammography systems (Scanner A on top-left vs. Scanner  B on bottom-left) as well as across medical institutions (Institution 1 on top-right v/s Institution 2 on bottom-right); (b) Summary of dataset composition, acquisition systems, and size. }\label{fig:intensity_histogram}
\end{figure}

Some recent works \cite{Roth_2020,yu2020salvaging} have shown that creation of site-specific models can potentially recover performance loss from catastrophic forgetting. However, having individualized models is more complicated to both train and deploy than a single, robust model is. Other approaches specifically handle catastrophic forgetting by constraining gradient updates to model parameters. For example, Elastic Weight Consolidation (EWC) \cite{kirkpatrick2017overcoming} updates model parameters proportionally to the inverse of each parameter's importance with respect to the original training dataset ($$\ie$$ the magnitude of updates for more important parameters is small). Zeng et al.~\cite{zeng2019continual} claim EWC to be ineffective in retaining performance on the original task and instead propose Orthogonal Weight Modification (OWM) wherein model parameters are updated in a direction orthogonal to the subspace spanned by the inputs of the model. However, a major limitation of OWM in medical applications is that OWM requires the underlying model to first perform feature extraction from the combined dataset before training a multi-layered perceptron on top of the extracted features. This is usually infeasible in our problem setting where datasets between multiple, medical institutions cannot be combined at a central location due to infrastructural as well as patient privacy issues. Moreover, both EWC and OWM have thus far been evaluated mainly in the context of \textit{continuous learning}, where the end goal is the sequential learning of separate tasks. This is different from our problem of domain expansion, where the goal is to learn a given task from data of a new domain without forgetting the original domain.

Unlike past works, we focus on the development of techniques to address catastrophic forgetting in a more \textit{realistic setting}. We consider a real-world clinical application of mammographic breast density assessment, which is routinely used to assess breast cancer risk to decrease the chances of breast cancer mortality \cite{tabar2001beyond,razzaghi2012mammographic}. Specifically, the identification of patients with dense breast tissue warrants additional monitoring, such as supplemental ultrasound or magnetic resonance imaging. The current criteria for mammographic breast density classification is based on the Breast Imaging Reporting and Data System (BI-RADS), which divides breast density into four distinct categories: fatty, scattered, heterogeneously dense, and extremely dense \cite{liberman1998breast}. In practice, however, BI-RADS is highly subjective and results in high inter-rater variability, which may confer undue patient anxiety and unnecessary, supplemental screening examinations \cite{sprague2016variation}. As such, there has been interest in developing automated approaches for assessment of mammographic density.

Most previous work that developed DL algorithms for breast density assessment have only focused on a single hospital/institution with a single digital mammography system \cite{lehman2019mammographic,mohamed2018deep}. A major hurdle for large-scale clinical deployment of a deep learning based breast density assessment tool is the poor generalizability across different hospitals/institutions and scanner types owing to inherent variability in patient demographics, disease prevalence, and imaging acquisition techniques \cite{zech2018variable,maartensson2020reliability,chang2020multi}. Our study addresses heterogeneity in digital mammography systems across different institutions (as depicted in Fig. \ref{fig:intensity_histogram}a) that arises from variability in x-ray tube targets, filters, digital detector technology, and control of automatic exposure \cite{keavey2012comparison}. 

We investigate domain expansion techniques across different digital mammography systems (Senographe as Scanner A and SenoScan as Scanner B) as well as across institutions (DMIST as Institution 1 and MGH as Institution 2)\footnote[2]{DMIST or digital mammographic imaging screening trial comprises of 33 institutions acquired from 5 digital mammography systems; MGH or Massachusetts General Hospital is a single institution dataset acquired from a single digital mammography system. See Section \ref{subsec:Method_Dataset} in Methods for additional details.} with an objective to mitigate catastrophic forgetting. Dataset composition and size details are enlisted in Fig. \ref{fig:intensity_histogram}b.
For simplicity, the original domain (Scanner A and Institution 1), is referred to as \textit{Dataset O} while the target domain (Scanner B and Institution 2) is referred as \textit{Dataset T}. 
The key contributions are as follows: 
\begin{itemize}
    \item We propose a simple yet effective technique to mitigate catastrophic forgetting by utilizing global BN statistics\footnote[3]{In this work, global BN statistics of a particular dataset refers to the running mean \& standard deviation of BN layers computed when training on that dataset. Typically, while batch statistics (batch mean \& batch standard deviation) are used during training, the global BN statistics are used only during inference. See Section \ref{subsec:Method_GBNS} in Methods for further details.} of Dataset O instead of Dataset T when fine-tuning on T.
    \item We demonstrate the efficacy of this technique under two different scenarios: first when restricting fine-tuning to only BN layers (motivated by \cite{karani2018lifelong}) and second when fine-tuning using all the layers. 
    \item We demonstrate how a commonly-used continuous learning algorithm (EWC \cite{kirkpatrick2017overcoming}) fails for large, real-world datasets and further highlight how augmenting EWC with our technique not only improves domain expansion over T but also mitigates catastrophic forgetting on O.
    \item Lastly, we provide a theoretical justification for why using global BN statistics of Dataset O instead of Dataset T better mitigates catastrophic forgetting.
\end{itemize}

%% file: Results.tex
\section{Results}
\begin{figure}[H]
\centering
  \subfloat[]{\includegraphics[width=0.5\textwidth]{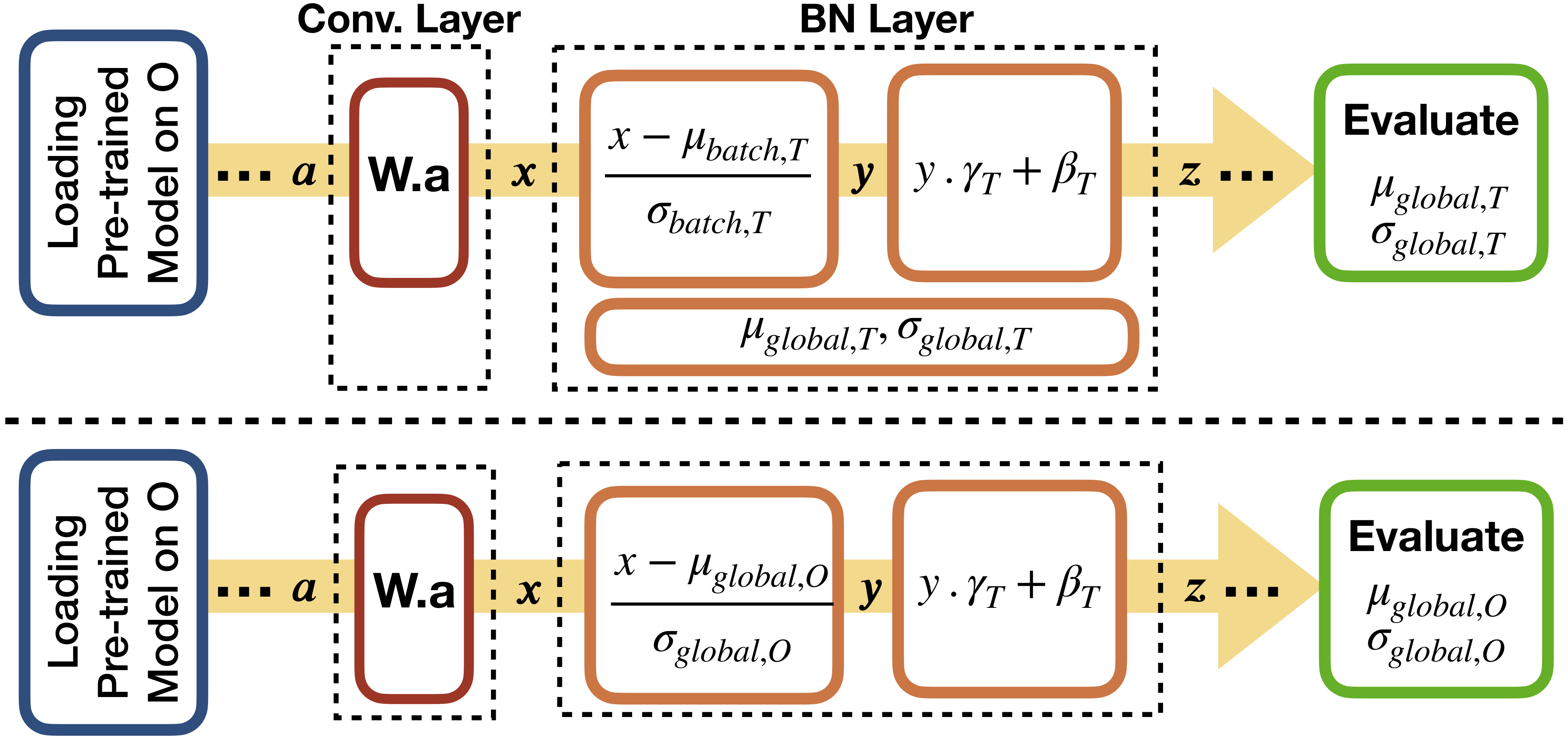}\label{fig:batchnorm_figure}}
  \hfill
  \subfloat[]{\includegraphics[width=0.5\textwidth]{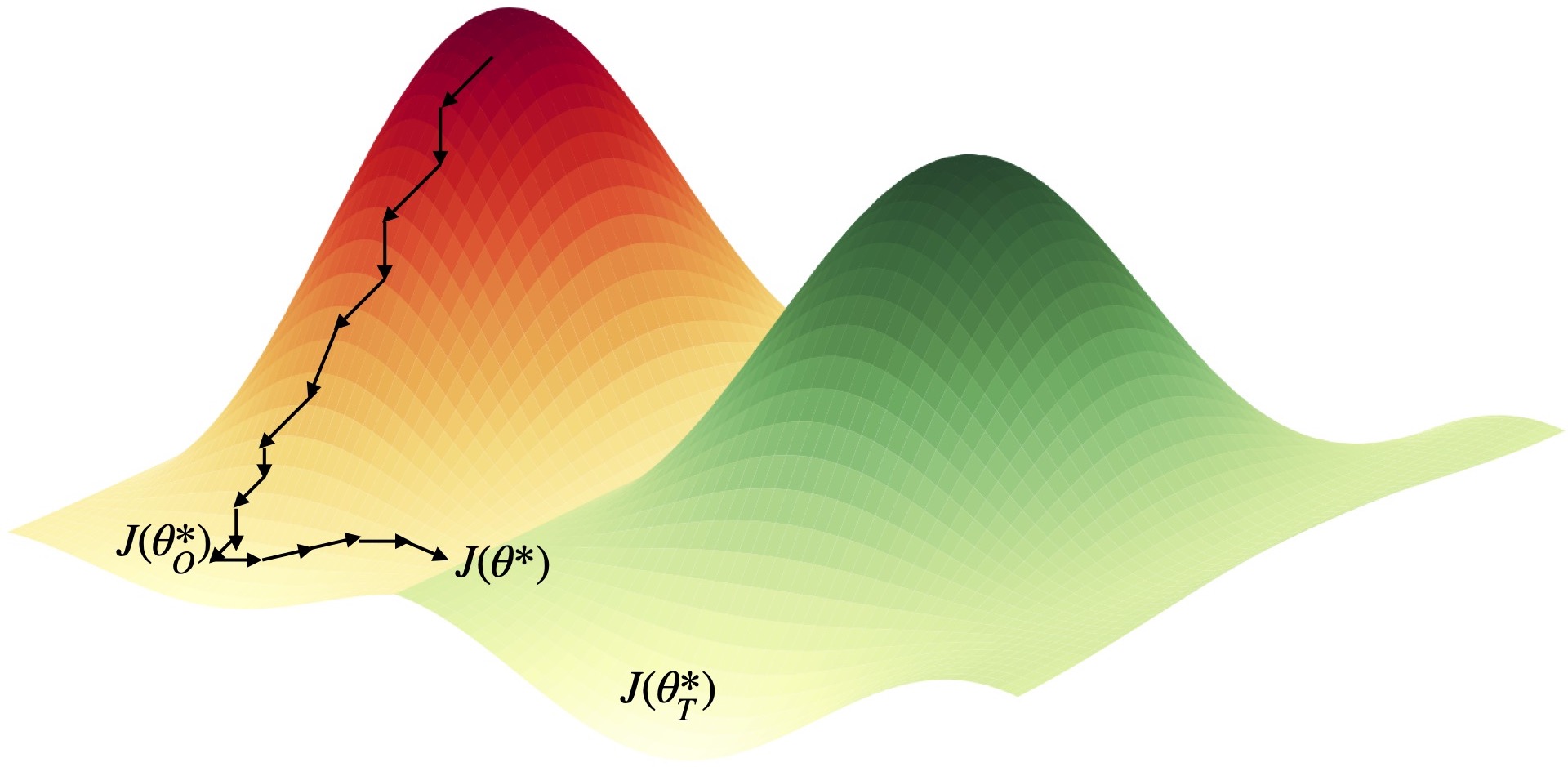}\label{fig:loss_landscape}}
  
\caption{(a) The two different BN approaches for fine-tuning on Dataset T: (Top) Using global BN statistics of Dataset T (running mean $\mu_{global,T}$ and running standard deviation $\sigma_{global, T}$ are computed during fine-tuning process); (Bottom) Using global statistics of Dataset O (running mean $\mu_{global,O}$ and running standard deviation $\sigma_{global,O}$ computed when originally training the model on O). (b) A schematic of EWC trajectory with the loss landscape in red corresponding to Dataset O and the one in green to Dataset T. The black line represents the evolution of loss of the fully trained baseline model on Dataset O when fine-tuned on Dataset T using EWC. The loss trajectory culminates in close proximity to both global minimas, resulting in high performance for both datasets.}
\label{fig:figure_ewc_batchnorm}
\end{figure}

With the objective of fine-tuning our breast density deep learning model on Dataset T and simultaneously mitigating catastrophic forgetting on Dataset O, we experiment with two broad sets of approaches: 
\begin{enumerate}
    \item Evaluation with global BN statistics of Dataset T (Fig. \ref{fig:batchnorm_figure} Top) vs.\ Dataset O (Fig. \ref{fig:batchnorm_figure} Bottom): both when fine-tuning with only BN layers\footnote[4]{We freeze all non-BN layers, allowing only BN parameters to train in this case.} and when fine-tuning with all layers.
    \item Evaluation after incorporating EWC (Fig. \ref{fig:loss_landscape}) using global BN statistics of Dataset O, again when fine-tuning with only BN layers and when fine-tuning with all layers. Kindly refer to Section \ref{subsec:Method_EWC} in Methods for additional details.
\end{enumerate}
 
\subsection{Baseline Experiments}
\begin{figure}[H]
\centering
  \subfloat[]{\includegraphics[width=0.75\columnwidth]{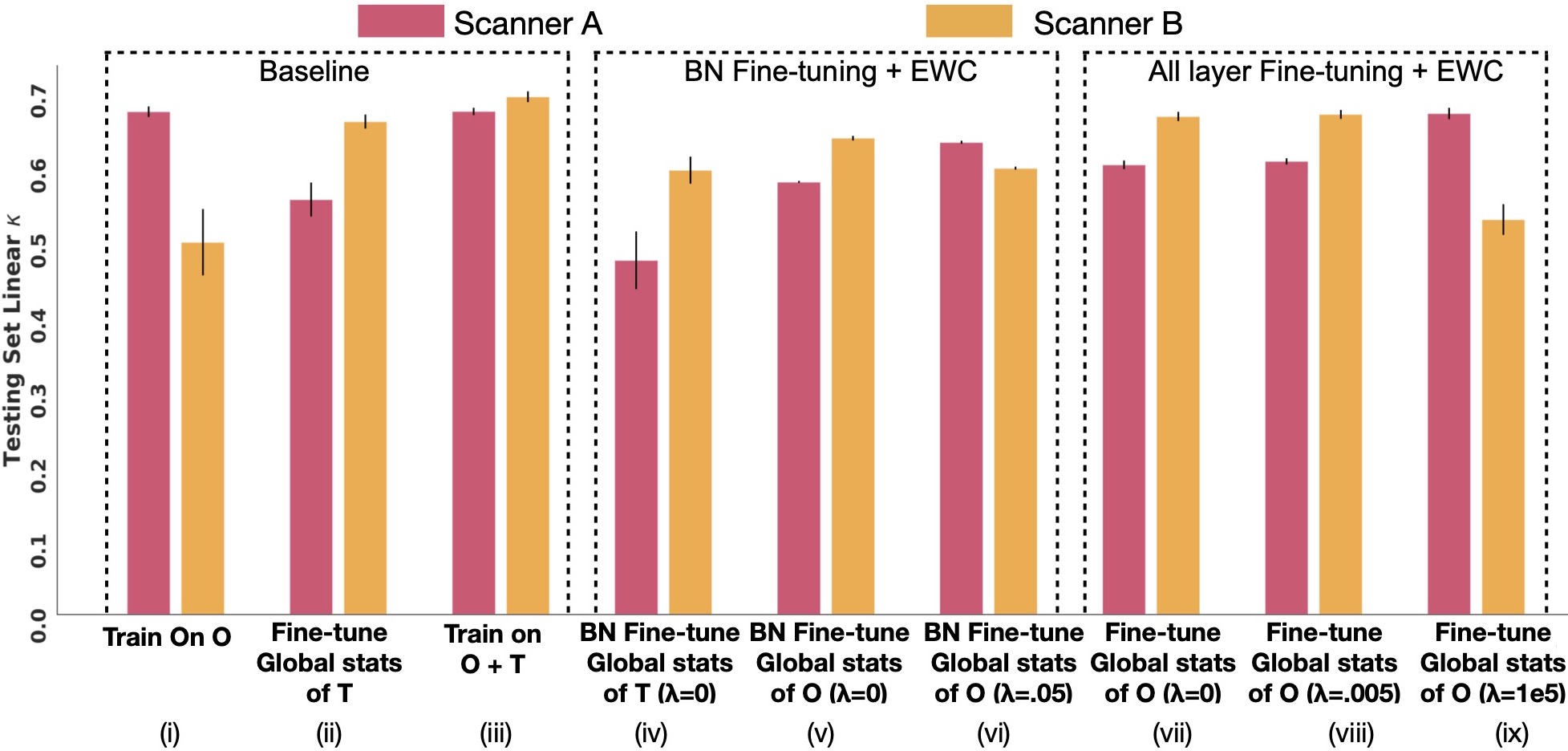}\label{fig:barplot1}}
  \hfill
   \subfloat[]{\includegraphics[width=0.75\columnwidth]{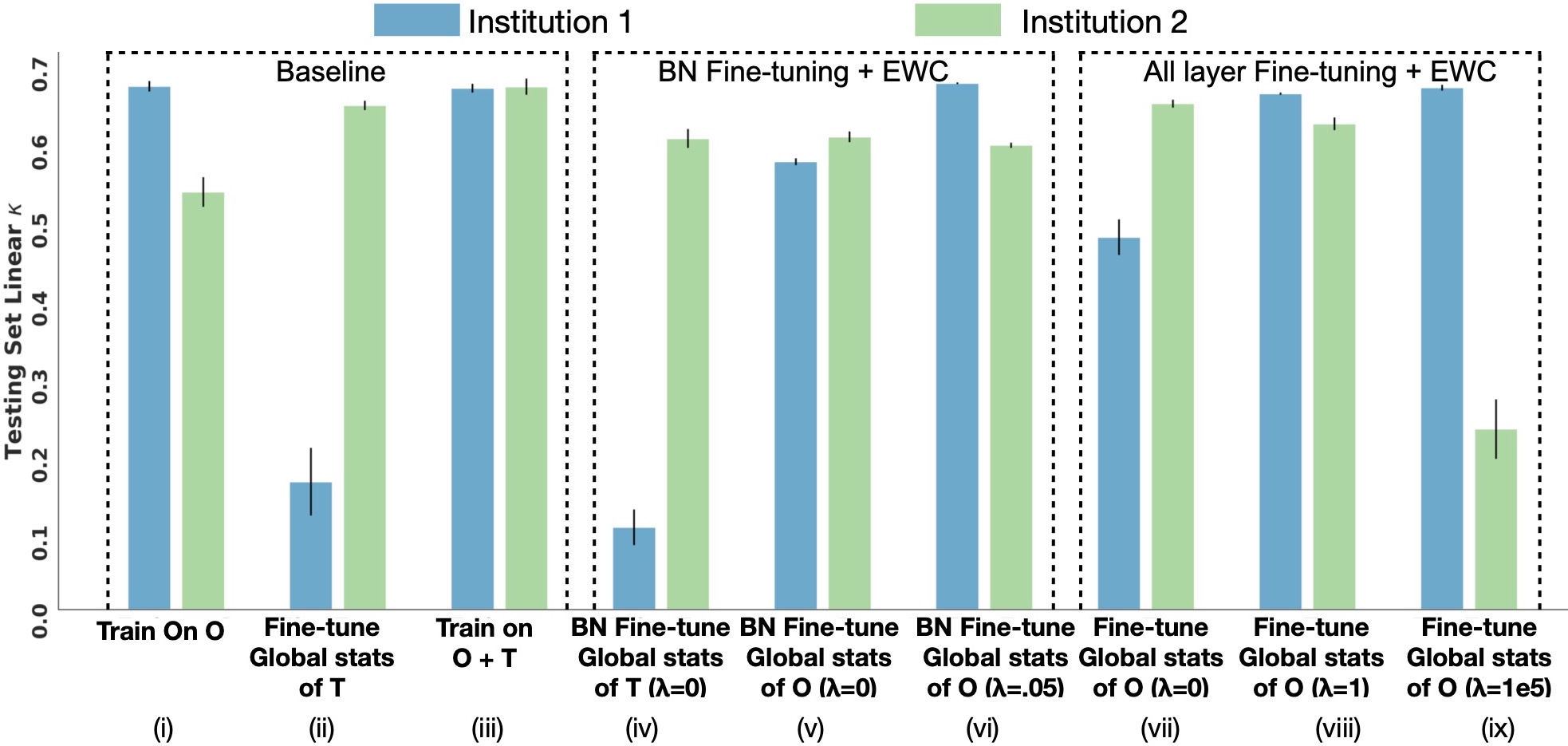}\label{fig:barplot2}}

\caption{Results for domain expansions across (a) digital mammography systems (Scanner A to Scanner B) and (b) institutions (Institution 1 to Institution 2). The different panels for each domain expansion type correspond to experiments for 1) Baseline models, 2) Only BN-layer fine-tuning with and without EWC\cite{kirkpatrick2017overcoming} and 3) All layer Fine-tuning with and without EWC\cite{kirkpatrick2017overcoming}.The parameter $\lambda$ is used as a trade off between the relative importance of performance over Datasets O and T (higher the $\lambda$ parameter, more closer will the performance of the fine-tuned model be to the model originally trained on Dataset O)}
\label{fig:barplot_combined}
\end{figure}

As a baseline, we performed experiments under three constructs: 1) training solely on Dataset O, 2) training on Dataset O and fine-tuning on Dataset T, and 3) training on the combined datasets O and T. 
 As shown in Fig. \ref{fig:barplot1}, \ref{fig:barplot2} (Panel ``Baseline''), starting from Dataset O (Scanner A or Institution 1) and moving onto Dataset T (Scanner B or Institution 2), we observe that models exclusively trained on Dataset O do not generalize well on Dataset T (Fig. \ref{fig:barplot1}, \ref{fig:barplot2} i). However, a model originally trained on O, when fine-tuned on T (Fig. \ref{fig:barplot1}, \ref{fig:barplot2} ii), abruptly forgets the information it learnt on O ($p < 0.01$ for both domain expansion across scanner types and institutions). Only when the model is trained collectively on Dataset O and T (Fig. \ref{fig:barplot1}, \ref{fig:barplot2} iii) does it achieve high performance on both domains with a performance of $\kappa$: 0.67 on Scanner A, $\kappa$: 0.69 on Scanner B and $\kappa$: 0.67 on Institution 1, $\kappa$: 0.67 on Institution 2. 



 

\subsection{Global BN statistics of target dataset (T) vs.\ original dataset (O)}

\subsubsection{Fine-tuning BN layers:}
With the intent of aligning the data distribution of Dataset T with Dataset O, we run several experiments fine-tuning only the BN layers for Dataset T while freezing all the non-BN layers. We start with the baseline model previously trained on Dataset O. This model is fine-tuned under two distinct constructs. In the first construct (traditional method as show in Fig. \ref{fig:batchnorm_figure} Top), this partially frozen model is fine-tuned using the batch statistics of Dataset T (batch mean $\mu_{batch,T}$ and batch standard deviation $\sigma_{batch,T}$) while its global BN statistics (running mean $\mu_{global,T}$ and running standard deviation $\sigma_{global, T}$) are concurrently calculated during the training process. The global BN statistics are then used to evaluate both datasets (Fig. \ref{fig:barplot1}, \ref{fig:barplot2} iv). In the second construct (as shown in Fig. \ref{fig:batchnorm_figure} Bottom), the global BN statistics of Dataset O (running mean $\mu_{global,O}$ and running standard deviation $\sigma_{global,O}$) are used for fine-tuning this model on Dataset T and again used for evaluation on both the datasets (Fig. \ref{fig:barplot1}, \ref{fig:barplot2} v).

From Fig. \ref{fig:barplot1} (Panel ``BN Fine-Tuning + EWC''), we observe that when the global BN statistics of Scanner B (T) are used (Fig. \ref{fig:barplot1} iv), the model undergoes a large reduction ($p < 0.01$) in the performance on Scanner A (O). Moreover, the performance on Scanner B (T) is lower ($p < 0.01$) compared to the performance when fine-tuned with all the layers (Fig. \ref{fig:barplot1} ii). In contrast, when the model is fine-tuned using the global BN statistics of Scanner A (O) (Fig. \ref{fig:barplot1} v), we see a recovery ($p < 0.01$) in the performance on Scanner A (O) as well as an increase in the performance on Scanner B (T) from the baseline model before fine-tuning ($p < 0.01$) (Fig. \ref{fig:barplot1} i). Similar results are obtained for Institution 1 (O) and Institution 2 (T). From these results, we can conclude that fine-tuning BN layers using the global BN statistics of Dataset O confers a performance advantage over fine-tuning using the global BN statistics of Dataset T. Moreover, although the performance on the target domain improves, fine-tuning with BN layers only partially mitigates catastrophic forgetting and therefore does not represent successful domain expansion.

\subsubsection{Fine-tuning all layers:}
Starting with the same baseline model specification as above, we again experiment with fine-tuning all layers of the model including the BN layers to utilize the full capacity of the model for learning new knowledge from Dataset T. Similar to the above setup, this model is fine-tuned for all layers over Dataset T and evaluated first using the global BN statistics calculated while training on Dataset T (Fig. \ref{fig:barplot1}, \ref{fig:barplot2} ii) and second using the global BN statistics of Dataset O (Fig. \ref{fig:barplot1}, \ref{fig:barplot2} vii). 
When a model trained on Dataset O is fine-tuned for all layers on Dataset T using the global BN statistics of Dataset T (Fig. \ref{fig:barplot1}, \ref{fig:barplot2} ii), we again observe that performance on Dataset O degrades ($p< 0.01$ for both Scanner A and Institution 1), the canonical presentation of catastrophic forgetting. Evaluating with global BN statistics of Dataset O (Fig. \ref{fig:barplot1}, \ref{fig:barplot2} vii) attenuates the performance loss ($p < 0.01$). In this construct, the model is able to perform well on the target domain (Scanner B and Institution 2) irrespective of the choice of global statistics used.

\subsection{Incorporation of Elastic Weight Consolidation}

\subsubsection{Fine-tuning BN Layers with EWC:}
\begin{figure}[H]
\centering
\includegraphics[clip, width=\columnwidth]{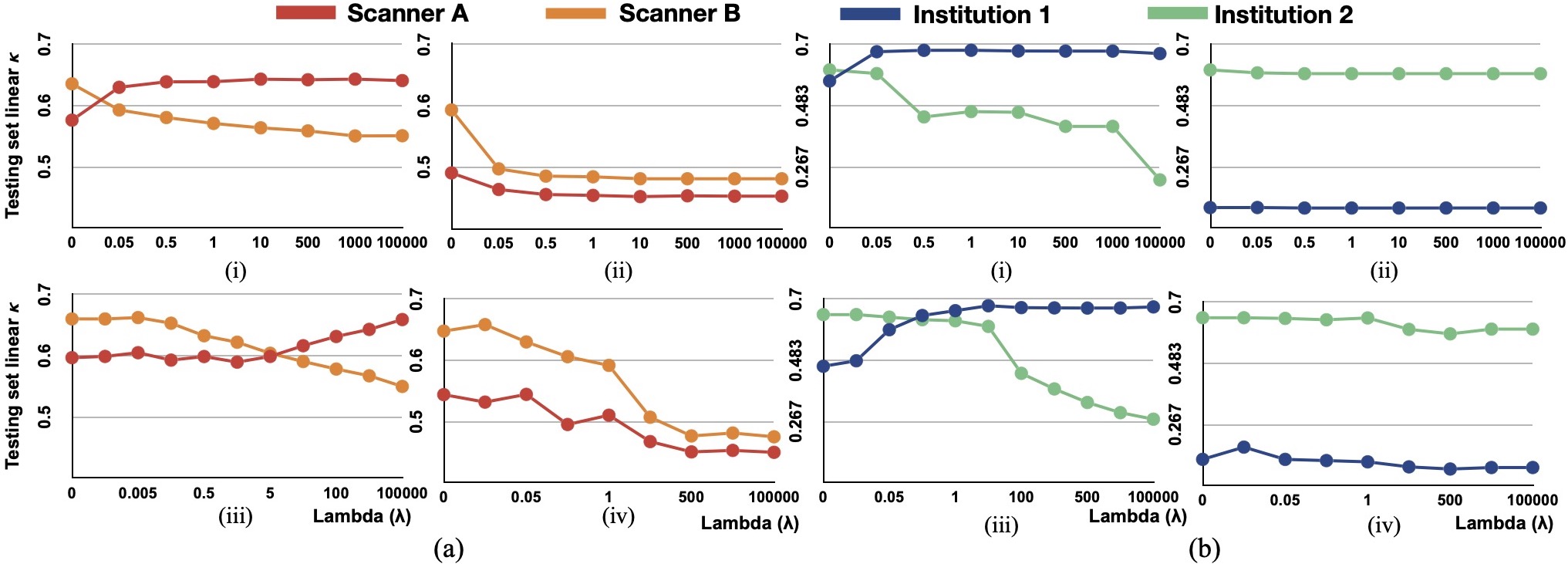}
\caption{Model performance while varying $\lambda$ when fine-tuning only BN layers (top row: (i),(ii)) vs.\ all layers (bottom row: (iii),(iv)) with EWC\cite{kirkpatrick2017overcoming} on (a) Scanner A and Scanner B using global BN statistics of Scanner A (left) vs.\ Scanner B (right) ; (b) Institution 1 and Institution 2 using global BN statistics of Institution 1 (left) vs.\ Institution 2 (right). Scanner A and Institution 1 being the original dataset O while Scanner B and Institution 2 being the target dataset T.}\label{fig:kappa_line_plot}
\end{figure}

We  explore the influence of EWC on the performance of the original and the target datasets when only the BN layers are fine-tuned using EWC (Fig. \ref{fig:barplot1}, \ref{fig:barplot2} vi). Performance on both datasets is evaluated after fine-tuning on Dataset T while varying the importance parameter ($\lambda$). After experimenting with a wide range of $\lambda$ values, we observe that when using the global BN statistics of Dataset O, at $\lambda=0.05$, we achieve maximum performance on both Dataset O and Dataset T (Fig. \ref{fig:kappa_line_plot}(a), \ref{fig:kappa_line_plot}(b) i). This value is optimal for both types of domain expansion: from Scanner A (O) to Scanner B (T) and from Institution 1(O) to Institution 2 (T). The performance at $\lambda=0.05$ on Scanner A (O) is $\kappa$: 0.63 and on Scanner B (T) is $\kappa$: 0.60. At the same $\lambda$, the performance on Institution 1 (O) and Institution 2 (T) is $\kappa$: 0.67 and $\kappa$: 0.59 respectively. 

\subsubsection{Fine-tuning all Layers with EWC:}
To further mitigate catastrophic forgetting on Dataset O and achieve peak performance on Dataset T, we allow fine-tuning of all layers while also incorporating EWC in the loss function. Performance on both datasets is evaluated after fine-tuning on Dataset T with varying importance parameters ($\lambda$). During fine-tuning with the global BN statistics of Dataset O, with increasing $\lambda$ (Fig. \ref{fig:kappa_line_plot}(a), \ref{fig:kappa_line_plot}(b) iii), performance on Dataset O consistently improves (red or blue curve in Fig. \ref{fig:kappa_line_plot}(a), \ref{fig:kappa_line_plot}(b) iii).
On the other hand, for the target dataset T (Scanner B or Institution 2), we see a degradation in performance with increasing $\lambda$ (orange or green curve in Fig. \ref{fig:kappa_line_plot}(a), \ref{fig:kappa_line_plot}(b) iii).  
When fine-tuning with the global BN statistics of Dataset O for very high $\lambda$ values ($\lambda=1e+5$), catastrophic forgetting for Dataset O is mostly mitigated, with a performance of $\kappa$: 0.67 and $\kappa$: 0.67 on Scanner A and Institution 1 respectively (Fig. \ref{fig:barplot1}, \ref{fig:barplot2} ix). However, very high $\lambda$ values prevent the model from learning the features of the target domain, and hence perform poorly on Scanner B ($\kappa$: 0.53) and Institution 2 ($\kappa$: 0.23) as seen in Fig. \ref{fig:barplot1}, \ref{fig:barplot2} ix. 

Overall, for domain expansion from Scanner A (O) to Scanner B (T) using the global BN statistics of Scanner A (O), the best performance on both Scanner A ($\kappa$: 0.60) and Scanner B ($\kappa$: 0.67) is obtained at $\lambda=0.005$ (Fig. \ref{fig:barplot1} viii). For domain expansion from Institution 1 (O) to Institution 2 (T) using the global BN statistics of Institution 1 (O), the best performance on both Institution 1 ($\kappa$: 0.66)and Institution 2 ($\kappa$: 0.62) occurs at $\lambda=1$ (Fig. \ref{fig:barplot2} viii). As expected, at the $\lambda$ which gives highest performance on both Datasets, fine-tuning BN layers rather than all layers (red and blue curves in Fig. \ref{fig:kappa_line_plot}(a), \ref{fig:kappa_line_plot}(b) when comparing i vs.\ iii) gives better performance on Dataset O (p $<$ 0.01 for both datasets). Conversely, fine-tuning all layers rather than just BN layers (orange and green curves in Fig. \ref{fig:kappa_line_plot}(a), \ref{fig:kappa_line_plot}(b) when comparing i vs.\ iii results in higher performance on Dataset T (p $<$ 0.01 for both datasets).  For both BN-only and all Layers fine-tuning, using Elastic Weight Consolidation (EWC) in conjunction with global BN statistics of the target dataset T,  is ineffective for mitigating catastrophic forgetting as compared to when using EWC with the global BN statistics of Dataset O (Fig \ref{fig:kappa_line_plot}(a), \ref{fig:kappa_line_plot}(b) comparing red or blue curves for left vs.\ right plots) further substantiating the importance of using global BN statistics of original dataset O. Thus, in summary, domain expansion is optimally performed by using both, the global BN statistic of Dataset O (to effectively mitigate catastrophic forgetting) and all Layers fine-tuning together with EWC (to attain peak perforamance on Dataset T).

%% file: Discussion.tex
\section{Discussion}

Continuous learning and multi-task learning have been an active area of research in the machine learning community. Although some techniques have been proposed \cite{kirkpatrick2017overcoming,zenke2017continual,farajtabar2020orthogonal,zeng2019continual}, they work well on simpler benchmark computer vision datasets (\eg Permuted MNIST \cite{goodfellow2013empirical}) and network architectures with few layers (2-3 densely connected layers), they often fail (see Section~\ref{subsec:SoTA} in the supplementary materials) when using deeper architectures with more complex layers and functions such as Dropout, ReLU activation, or BN.

Normalization layers have become a ubiquitous technique for training deep neural networks. Batch Normalization (BN) specifically, normalizes the output from a layer using the mean and standard deviation of the input batch, effectively allows training with higher learning rates and usually improves convergence to a better local optimum \cite{ioffe2015batch}. While most existing work focuses on the influence of BN when training models on a \textit{single institution dataset}, few \cite{karani2018lifelong} have actually tried to adapt the BN layers for training these models sequentially over multiple datasets with varied scanner types and protocols. Karani et. al. \cite{karani2018lifelong} froze the convolutional layers and trained only the batch norm layers independently for each dataset. Although this approach allows the model to retain performance on the previously trained distribution and task, it restricts model capacity to learn newer tasks or domains. Moreover, every institution would need to have separate BN parameters, which defeats the purpose of \textit{one model for both the original and target datasets}. In our approach, we allow training of all layers on the target dataset using EWC\cite{kirkpatrick2017overcoming}, while simultaneously fixing BN statistics to the global BN statistics from the original dataset O, thus preventing catastrophic forgetting on O. Contrary to Karani et. al. \cite{karani2018lifelong}, we envision one universal model for both original and target distributions involved during domain expansion.

Plasticity is the capacity of a network to adapt to new environments while stability ensures retention of previously learned knowledge. 
Setting the appropriate trade-off between stability and plasticity of a neural network is critical, not only to avoid forgetting but also to learn new tasks quickly. Catastrophic forgetting occurs when a network is overly plastic and not sufficiently stable; \ie the network is able to quickly acquire new tasks or modalities but does not retain previously learned tasks and modalities. \cite{mirzadeh2020dropout} studied this behaviour in the context of dropout layers and showed how dropout can overcome catastrophic forgetting on a previous task when the same model is trained on a new task. Specifically, they showed that  multi-layered perceptrons trained with dropout regularization have higher stability and lower variance in the output after ReLU activation function. In other words, activation which are purely active or inactive (values 0 or 1) remain untouched (ensuring stability on previously learned tasks) and only the semi-active neurons are turned off or on, thus promoting plasticity for newer tasks. EWC \cite{kirkpatrick2017overcoming} works by a similar principle, \ie ensuring stability by constraining those weights which are most important for previous learned tasks, while still retaining enough plasticity by allowing other weights to be trained for newer tasks.

\begin{figure}[H]
\centering
\includegraphics[clip, width=\columnwidth]{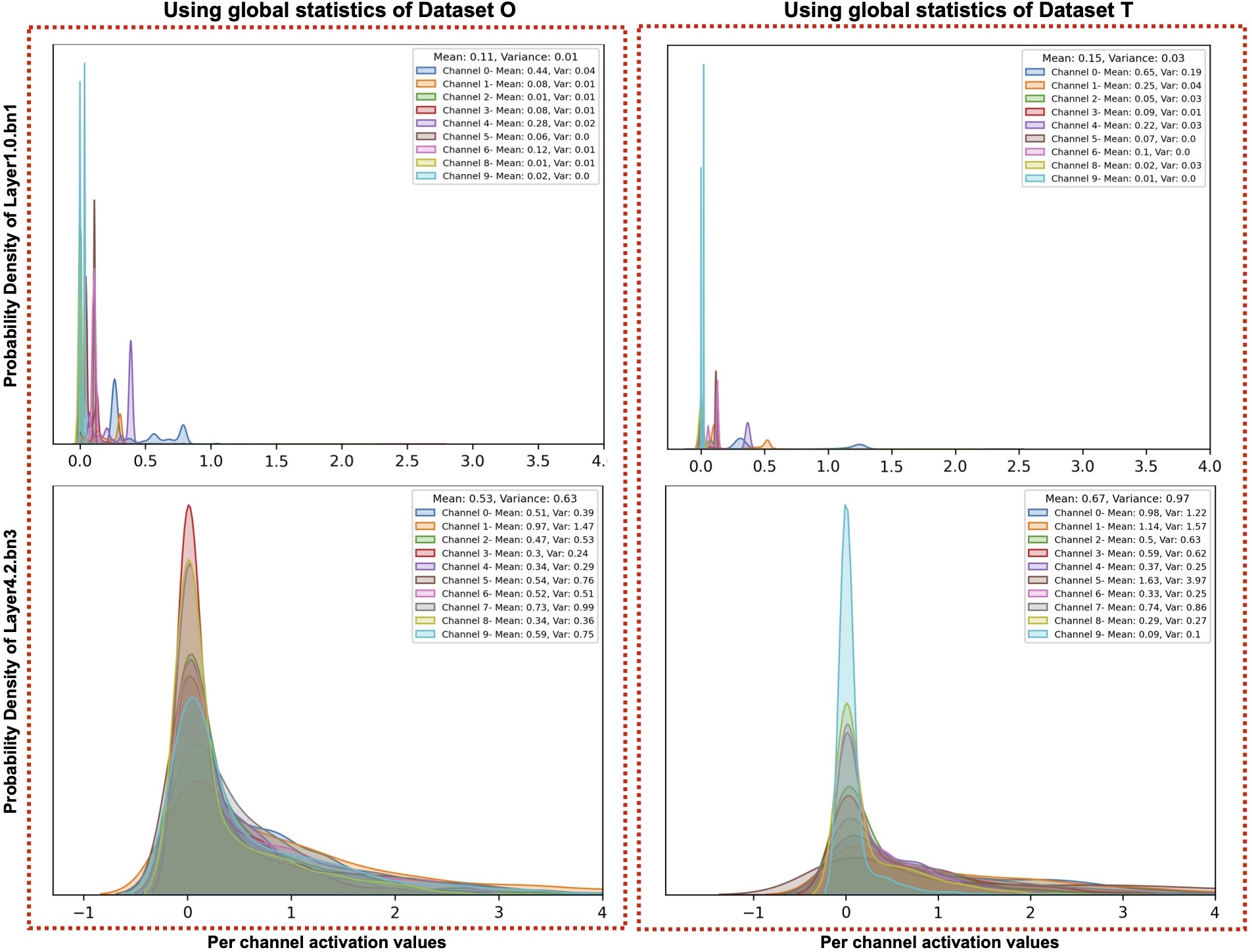}
\caption{Probability Density of per-channel activation values computed over a batch from original Dataset O after fine-tuning on target Dataset T using global statistics of O (left) and T (right) for an initial layer Layer1.0.bn1 (top) and a deeper layer Layer4.2.bn3 (bottom) of the Resnet-50 architecture. Here layer1.0.bn1 represents the first BN layer in the Bottleneck layer of ResNet50 architecture while layer4.2.bn3 represents the last BN layer before the final classification layer}\label{fig:MeanActivations}
\end{figure}

In order to better understand the influence of global BN statistics on overall stability of a network fine-tuned over Dataset T, we plot the distribution of activation values from different BN layer outputs in the network using samples from Dataset O (Fig. \ref{fig:MeanActivations}). Using global BN statistics of O (Fig. \ref{fig:MeanActivations} left), we empirically see that deeper layers of the network (Fig. \ref{fig:MeanActivations} top vs.\ bottom),  have a lower activation variance than when using global BN statistics of T (Fig \ref{fig:MeanActivations} right). Notably, while the mean variance across different channels for the model fine-tuned using global BN statistics of O increases from 0.01 to 0.63, the model fine-tuned using the global BN statistics of T shows a much greater increase from 0.03 to 0.97. For a detailed, theoretical proof corresponding to this observation, we refer the reader to Section \ref{subsec:proof} in Method. 

\begin{figure}[H]
\centering
\includegraphics[clip, width=\columnwidth]{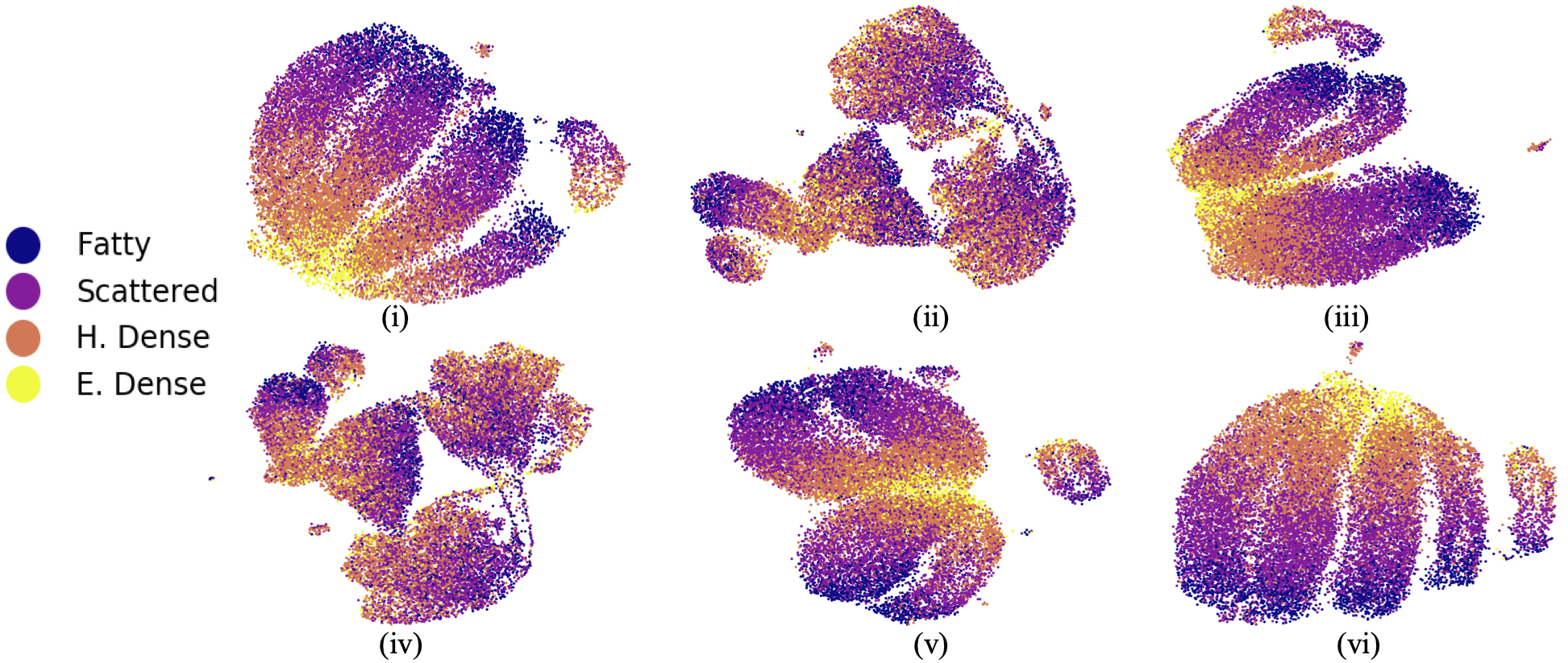}
\caption{UMAPs for original Dataset O when (i) Model is trained just on Dataset O; When fine-tuning only BN layers with global BN statistics of (ii) T; (iii) O; Fine-tuning all layers with global BN statistics of (iv) T; (v) O; (vi) O after incorporating EWC (with optimal $\lambda$ value)}\label{fig:umaps}
\end{figure}

Inspired by \cite{mirzadeh2020dropout}, we further hypothesize that as we move deeper into a network fine-tuned over Dataset T, the lower variance (from using global BN statistics of O instead of T) promotes greater stability of the network and would therefore lead to lower plasticity and less catastrophic forgetting. To further investigate the influence of global BN statistics on catastrophic forgetting, we visualize the UMAPs \cite{mcinnes2018umap} of features computed over Dataset O from the various models in our experiments. Fig. \ref{fig:umaps}(i) depicts UMAPs of sample features for the model trained originally on Dataset O. We observe clear separation for different classes, aligned along a particular direction in low-dimensional space that likely corresponds with high overall performance. In Fig. \ref{fig:umaps}(ii) or \ref{fig:umaps}(iv) (representing models fine-tuned with only BN layers and fine-tuned with all layers respectively by using global BN statistics of T), we observe that UMAPs of features from Dataset O have classes which are highly entangled with each other, which correlate with lower, overall performance on O. However, for Fig. \ref{fig:umaps}(iii) or \ref{fig:umaps}(v) (representing models fine-tuned with only BN layers and fine-tuned with all layers, respectively, by using global BN statistics of O), we observe how different classes which show high separation results in avoiding any forgetting. This is also observed when fine tuning all layers after incorporating EWC and using global statistics of O (Fig \ref{fig:umaps} (vi) shows classes are well segregated and aligned in one direction). Overall, this illustrates how global BN statistics of O prevent large variance shifts in the deeper layers, which may avoid any serious entanglement of features in deeper layers, and ultimately results in higher stability of the network to mitigate catastrophic forgetting. 

%% file: Conclusion.tex
\section{Conclusion}

In this study, we develop a simple yet novel approach to avoid catastrophic forgetting while fine-tuning deep learning models on new datasets, enabling continuous learning and refinement. Catastrophic forgetting has major implications for clinical deployment of deep learning models, where generalizability remains a major hurdle to large-scale deployment. Specifically, we demonstrate that relying on global BN statistics of the original Dataset O when fine-tuning on the target Dataset T alleviates forgetting to a large extent, though the performance achieved on Dataset T isn't optimal. However, when the above technique is used in conjunction with EWC, we are not only able to completely mitigate catastrophic forgetting on Dataset O but also achieve peak performance on Dataset T. We further observe that fine-tuning only BN layers (similar to \cite{karani2018lifelong}) is insufficient to achieve high performance on T. Rather, fine-tuning all layers is necessary to achieve high performance on both Dataset O and T.

There are still several limitations and possible future directions to this work. First, we only evaluate our approach in the context of domain expansion where a model trained on one dataset is fine tuned on a second dataset; future work can investigate the scalability of our approach when domain expansion is performed across a larger number of datasets. Furthermore, distributed federated learning has received considerable attention recently, primarily for training models over multi-institutional medical datasets, which are difficult to share due to both privacy and legal issues. We believe that applying our approach in the distributed learning contexts \cite{Roth_2020,sheller2020federated,chang2018distributed} to see if performance can be improved, particularly in scenarios where there is high heterogeneity across institutions \cite{balachandar2020accounting}, is another promising direction of our work.



%% file: Method.tex
\section{Methods}


  

\subsection{Global BN statistics of target dataset (T) vs.\ original dataset (O)}\label{subsec:Method_GBNS}
Batch normalization (BN) has become a popular technique to regulate shifts in the distribution of network activations during training \cite{ioffe2015batch}. BN normalizes each batch of input to the network layers by subtracting the batch mean and dividing by the batch standard deviation. Fig. \ref{fig:batchnorm_figure} (top row) illustrates pipeline for traditional fine-tuning on Dataset T starting from a pre-trained model on Dataset O. During fine tuning, output from the previous convolutional layer is passed to the BN layer, which normalizes the input $x_{1},x_{2}...x_{m}$ (where $m$ is the batch size):

\begin{equation} \label{eq:batchnorm}
 y_{i} = \gamma_{T}\frac{x_i-\mu_{batch,T}}{\sqrt{\sigma^{2}_{batch,T}+\epsilon}} + \beta_{T},
\end{equation}

\noindent where $y_{i}$ is the output after applying BN on input $x_{i}$, $\mu_{batch,T}$ and $\sigma^{2}_{batch,T}$ are the batch mean and batch variance of Dataset T, respectively, $\epsilon$ is a stability parameter, and $\gamma_{T}$ and $\beta_{T}$ are trainable parameters of BN layers when fine-tuned with Dataset T. 
During inference, BN uses the global BN statistics of Dataset T (running mean $\mu_{global,T}$ and the running standard deviation $\sigma_{global, T}$) which were calculated while training; specifically, at training iteration k, the running mean $\mu_{global,T}[k]$ and running variance $\sigma^{2}_{global,T}[k]$ for Dataset T are updated: 

 \begin{center}
    \begin{equation*}
    \mu_{global,T}[k] = \alpha  \mu_{global,T}[k-1] + (1-\alpha) * \mu_{batch,T}[k]
 \end{equation*} 
 \end{center}

\begin{center}
    \begin{equation*}
   \sigma^{2}_{global,T}[k] = \alpha \sigma^{2}_{global,T}[k-1] + (1-\alpha) * \sigma^{2}_{batch,T}[k]
 \end{equation*} 
 \end{center}
 
\noindent  where the default momentum parameter, $\alpha$, is set at 0.9. During inference, we replace $\mu_{batch,T}$ and $\sigma^{2}_{batch,T}$ in equation \ref{eq:batchnorm} with the global BN statistics of Dataset T computed during training, resulting in:
\begin{equation} \label{eq:0}
 y_{i} = \gamma_{T}\frac{x_i-\mu_{global,T}}{\sqrt{{\sigma^{2}_{global,T}}+\epsilon}} +\beta_{T}
 \end{equation}



\noindent In addition to the traditional approach to compute global BN statistics ($\mu_{global,T}$ and ${\sigma_{global,T}}$) fine-tuned on Dataset T and evaluated with the same global BN statistics for both datasets, we also conduct experiments to both fine-tune and evaluate using global BN statistics of Dataset O (as shown in bottom row of Fig \ref{fig:batchnorm_figure}). Thus, the BN (both for fine-tuning and inference) on an input batch can be represented by:

\begin{equation} \label{eq:BNO}
 y_{i} = \gamma_{T}\frac{x_i-\mu_{global,O}}{\sqrt{{\sigma^{2}_{global,O}}+\epsilon}} +\beta_{T}
\end{equation}
where $\mu_{global,O}$ and $\sigma_{global,O}$ are the global BN statistics of Dataset O (running mean and running standard deviation computed when training the model originally on Dataset O); $\gamma_{T}$ and $\beta_{T}$ denote the trainable parameters of BN layers when fine-tuning with Dataset T but using global BN statistics of O.

We experiment with fine-tuning the model using the above two global BN statistics techniques under the following two scenarios:
\begin{itemize}
    \item \textbf{Fine tuning only BN layers}: As shown in Fig \ref{fig:intensity_histogram}, the intensity distributions of Dataset O (Scanner A in red and Institution 1 in blue) differ considerably from the Dataset T (Scanner B in orange and Institution 2 in green), despite all being mammography screening datasets. With the intuition of aligning the data distribution of Dataset T with Dataset O, and retaining the original model performance for Dataset O, we first test run several experiments by fine-tuning only BN layers on Dataset T while freezing all other layers.
    \item \textbf{Fine-tuning all layers (BN and convolutional layers)}: With the primary objective of achieving the best performance on Dataset T while retaining the original performance on Dataset O, we further investigate the effects of fine-tuning all layers. Training using all layers aids in leveraging the full capacity of a model for fine-tuning on Dataset T.
\end{itemize}


\subsection{Incorporation of Elastic Weight Consolidation (EWC)}\label{subsec:Method_EWC}
While fine-tuning a model (previously trained on Dataset O) on Dataset T, EWC \cite{kirkpatrick2017overcoming} effectively constrains the updates of those weights which are most important for Dataset O, thus allowing the model to converge to a minimum close to global minima of both datasets T and O (as shown in Fig \ref{fig:loss_landscape} and ultimately prevents catastrophic forgetting for Dataset O.

We implement this constraint using a quadratic penalty on the change in model parameters. In order to identify the salient parameters for model trained on Dataset O, we first compute the empirical Fisher Information Matrix (FIM). \cite{Ly2017ATO} The Fisher information matrix is defined as the covariance of the score function used to calculate the quality of parameter estimation. Due to the complexity of the likelihood function, computing this expectation becomes intractable and hence an approximation called the empirical Fisher (F) is defined as 



\begin{equation} \label{eq:fisher_def}
F= \frac{1}{N}\sum_{i=1}^{N}\nabla_{\theta}\log p(x|\theta)\nabla_{\theta}\log p(x|\theta)^{'}
\end{equation}

This empirical Fisher matrix when combined with the quadratic penalty adds a constraint to the important parameters of model trained on Dataset O. The combined loss function $L(\theta)$ for elastic weight consolidation is given by 
\begin{equation}
    L(\theta) = L_{T}(\theta) + \sum_i \frac{\lambda}{2}F_{i}(\theta_{i}-\theta^{*}_{O,i})^{2}
\end{equation}
where $L_{T}(\theta)$ is the loss function for Dataset T, $\theta^{*}_{O}$ represents the optimal model parameters for Dataset O and $i$ in $\theta_{i}$ iterates over all the current model parameters which are being fine-tuned on Dataset T. The parameter $\lambda$ is used as a trade off between the relative importance of performance over Datasets O and T (the higher the $\lambda$ parameter, the more closer will be the performance of the fine-tuned model to the model originally trained on Dataset O) .

As a result, a constraint on some (not all) model parameters, allows the solution to stay in a low-error region which is optimal for both the datasets \cite{kirkpatrick2017overcoming}.
We experiment with fine-tuning the model using EWC together with global BN statistics of O under the following two conditions: fine-tuning only the BN layers, and fine-tuning all layers.

\subsection{Comparative analysis of variance shifts with global BN statistics of T and O}\label{subsec:proof}

From \cite{mirzadeh2020dropout}, we know that the variance of the penultimate layer's activations of a network directly impacts the stability of a network. In other words, the lower the variance of the activations of the latter layers, the smaller the extent of model's catastrophic forgetting is likely to be. In order to illustrate the influence of using the global BN statistics of T and O on the overall stability of a network, we compute the variance of the outputs of the BN layers of a model fully fine-tuned on Dataset T (without EWC),  given in equations \ref{eq:0} and \ref{eq:BNO}.
The variance of a layer's activations for the model trained using global BN statistics of T from equation \ref{eq:0} is given by:
\begin{equation} \label{eq:var_BNT}
Var[y_{i}] = Var\left[\gamma_{T}\frac{x_i-\mu_{global,T}}{\sqrt{{\sigma^{2}_{global,T}}+\epsilon}} +\beta_{T}\right]
\end{equation}
Assuming $\gamma_{T}$, $\beta_{T}$, $\mu_{global,T}$ and $\sigma^{2}_{global,T}$ to be constant at the time of inference and thereafter for the purpose of understanding catastrophic forgetting, Dataset O is used for inference:
 \begin{center}
    \begin{equation}
    Var[\beta_{T}] = 0\newline
    \end{equation} 
 \end{center}
\begin{center}
    \begin{equation}
    Var\left[\gamma_{T}\frac{x_i-\mu_{global,T}}{\sqrt{{\sigma^{2}_{global,T}}+\epsilon}}\right] = \frac{\gamma^{2}_{T}}{\sigma^{2}_{global,T}+\epsilon}Var(x_{i})
 \end{equation} 
 \end{center}
Hence, the final variance of output from BN layer of a model trained using global BN statistics of T is given by:
\begin{equation} \label{eq:var_BNT2}
Var[y_{i}] = C_{T} \times Var(x_{i})
\end{equation}
where $C_{T}$ is any constant. Next, the variance for the global BN statistics of O from equation \ref{eq:BNO} is given by:
\begin{equation} \label{eq:var_BNO1}
Var[y_{i}] = Var\left[\gamma_{T}\frac{x_i-\mu_{global,O}}{\sqrt{{\sigma^{2}_{global,O}}+\epsilon}} +\beta_{T}\right]
\end{equation}
Again assuming $\gamma_{T}$, $\beta_{T}$, $\mu_{global,O}$ and $\sigma^{2}_{global,O}$ to be constant at the time of inference and thereafter for the purpose of understanding catastrophic forgetting, Dataset O is used for inference:
\begin{center}
    \begin{equation}
    Var[\beta_{T}] = 0 
     \end{equation} 
 \end{center}
 \begin{center}
    \begin{equation}
    Var\left[\frac{x_i-\mu_{global,O}}{\sqrt{{\sigma^{2}_{global,O+\epsilon}}}}\right] \approx 1
 \end{equation} 
 \end{center}
 Hence, the final variance of output from BN layer of a model trained using global BN statistics of O is given by:
 \begin{equation} \label{eq:var_BNO2}
  Var[y_{i}] = \gamma^{2}_{T} = C_{T}
 \end{equation}

Thus while the variance in the case of the model trained using the global BN statistics of O (equation \ref{eq:var_BNO2}) is simply a constant (based on the scaling parameter of the corresponding BN layer), the variance in the case of model trained using the global BN statistics of T (equation \ref{eq:var_BNT2}) is a constant times the variance of the input to the BN layer. In other words, though we see some variance shift in both cases, the shift in case of global BN statistics of T (equation \ref{eq:var_BNT2}) keeps on increasing as we go deeper into the network (owing to its direct proportionality to input variance). This increase in the variance shift over the layers of a network, ultimately results in extreme distortions in the features computed from the penultimate layers (as seen in Fig. \ref{fig:umaps}) of a model trained using global BN statistics of T when tested on the original dataset O, thus leading to poor stability of the network and ultimately resulting in adverse catastrophic forgetting over O.

\subsection{Datasets and Preprocessing}\label{subsec:Method_Dataset}

Digital screening mammograms from 33 institutions were retrospectively obtained through the Digital Mammographic Imaging Screening Trial (DMIST), the details of which were previously published \cite{pisano2005diagnostic}. This study was approved by the Institutional Review Board (IRB) of the American College of Radiology Imaging Network (ACRIN), by the IRB and the Cancer Therapy Evaluation Program at the National Cancer Institute.
For DMIST, a total of 92 radiologists from the United States and Canada read the exams. Readers in the United States were all qualified interpreters of mammograms under federal law. Canadian readers met equivalent standards. Each site’s lead radiologist received training to read for DMIST and in turn trained the site’s other readers. DMIST images were previously de-identified for this study. For DMIST, 5 digital mammography systems were used: SenoScan (Fischer Medical), the Computed Radiography System for Mammography (Fuji Medical), the Senographe 2000D (General Electric Medical Systems), the Digital Mammography System (Hologic), and the Selenia Full Field Digital Mammography System (Hologic)\cite{pisano2005diagnostic}. The mammograms were saved in DICOM format with 4 different image data formats, corresponding to different digital-mammography systems or different versions of the same system: 12 bit Monochrome 1 (30.3\%), 12 bit Monochrome 2 (11.2\%), 14 bit Monochrome 1 (58.0\%), and 14-bit Monochrome 2 (0.5\%). 14-bit Monochrome 2 images were excluded to ensure that each image data format included in our study had adequate representation for training of our deep learning model. The final DMIST patient cohort consisted of 108,230 digital screening images from 21,759 patients. 

We also obtained digital screening mammograms from Massachusetts General Hospital (MGH) following IRB approval. Patients who had prior surgery or implants were excluded. For MGH, each image was read by a breast imaging radiologist as part of routine clinical practice. All mammograms were acquired using a Lorad Selenia mammography system (Hologic). The final MGH patient cohort consisted of 8,603 digital screening images from 1,856 patients. 

All images from DMIST and MGH were interpreted by a single radiologist from a pool of radiologists using the ACR BI-RADS breast density lexicon (Category A: fatty, Category B: scattered, Category B: heterogeneously dense, Category D: extremely dense)\cite{liberman1998breast}. The wide variations in intensity histograms between the four datasets is illustrated in Fig. \ref{fig:intensity_histogram}a.

The SenoScan (Scanner B, $n = 32928$ ), Senographe (Scanner A, $n = 59411$), DMIST (Institution 1, $n = 103890$), MGH (Institution 2, $n = 8603$) patient cohorts were split into training, validation, and testing sets in a 7:2:1 ratio, on a patient level. The training set was used to develop the model and the validation set was used to assess model performance during training to prevent overfitting. The test set was unseen until the model training was completed. The intensity of each image was scaled to be between 0 and 1 by dividing by the maximum value of the image format (4095 for 12 bit and 16383 for 14 bit). All Monochrome 1 images were inverted to make them equivalent to Monochrome 2. To ensure proper input size to the pre-trained neural network architectures, the images were resized to $224 \times 224 \times 3$.

 

\subsection{Training and Prediction}
For the baseline classification model, a Resnet50 architecture with ImageNet-pretrained weights\cite{he2016deep} was used. Cross entropy loss function and the Adam optimizer \cite{kingma2014adam} (lr = $10^{-6}$, $\beta_1 = 0.9$, $\beta_2 = 0.99$, $\epsilon = 10^{-7}$) were used across models. Batches were randomly sampled using a fixed batch size of 32 images. Early stopping with a patience of 20 epochs is used to prevent overfitting. Checkpoints were saved after each epoch based on the performance on the validation set and the model with highest validation accuracy is saved and reported. The training set is augmented in real time by means of random flips and rotations of at-most 45 $\degree$.

To combine predictions from all images across all mammography views from a given patient study into a patient-level assessment, the output probabilities for all corresponding images from a patient study are averaged. The averaged probabilities are then used to determine the predicted breast density class. All models are evaluated using Cohen's Kappa scores with linear weighting ($\kappa$). For reference, a $\kappa$ of 0.21-0.40, 0.41-0.60, and 0.61-0.80 represents fair, moderate, and substantial agreement, respectively \cite{kappa1977original}. A Wilcoxon signed-rank test at a significance level of $p = 0.05$ was used for statistical comparisons of model performance.



%% file: Acknowledgments
\section{Acknowledgements}

Research reported in this publication was supported by a training grant from the National Institute of Biomedical Imaging and Bioengineering (NIBIB) of the National Institutes of Health under award number 5T32EB1680 to K. Chang and J. B. Patel and by the National Cancer Institute (NCI) of the National Institutes of Health under Award Number F30CA239407 to K. Chang. The content is solely the responsibility of the authors and does not necessarily represent the official views of the National Institutes of Health.

This publication was supported from the Martinos Scholars fund to K. Hoebel. Its contents are solely the responsibility of the authors and do not necessarily represent the official views of the Martinos Scholars fund.

This study was supported by National Institutes of Health (NIH) grants U01CA154601, U24CA180927, and U24CA180918 to J. Kalpathy-Cramer, U01CA242879 to D. Rubin and J. Kalpathy-Cramer, and National Science Foundation (NSF) grant NSF1622542 to J. Kalpathy-Cramer. This research was carried out in whole or in part at the Athinoula A. Martinos Center for Biomedical Imaging at the Massachusetts General Hospital, using resources provided by the Center for Functional Neuroimaging Technologies, P41EB015896, a P41 Biotechnology Resource Grant supported by the National Institute of Biomedical Imaging and Bioengineering (NIBIB), National Institutes of Health.

Lastly, we would also like to thank ECOG-ACRIN for providing access to the DMIST dataset and to the ACR-DSI for valuable discussions. Original data collection for ACRIN 6652 (DMIST) was supported by NCI Cancer Imaging Program grants.

%% file: Supplement.tex
\section{Supplementary Material}

\subsection{Evaluation of Validation Loss}

\begin{figure}[H]
\centering
\includegraphics[clip, width=\columnwidth]{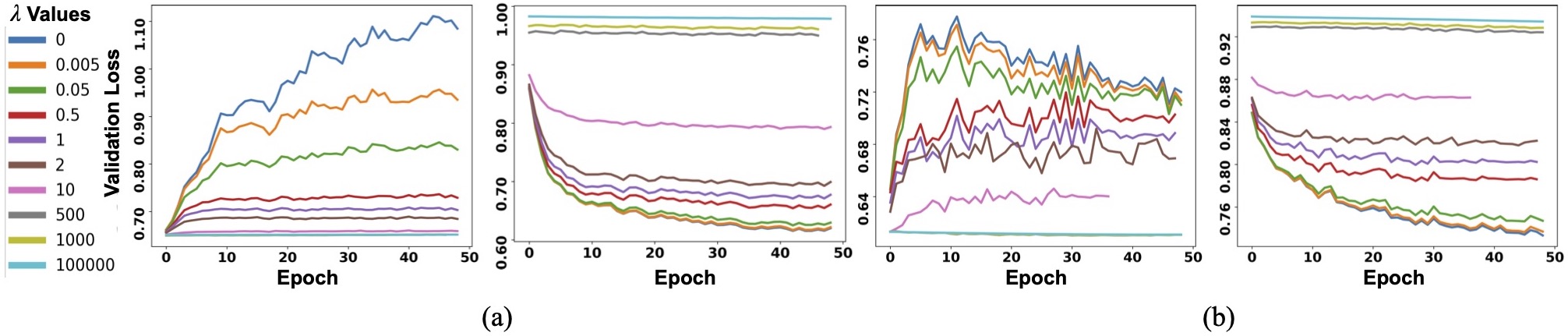}
\caption{Evolution of validation loss during All-layer fine-tuning using EWC for different $\lambda$ values for (a) Institution 1 (left) and Institution 2 (right); (b) Scanner A (left) and Scanner B (right).}\label{fig:lossplot_ewc}
\end{figure}
We visualize the evolution of the validation loss (Fig. \ref{fig:lossplot_ewc}) on both the original and target datasets with varying $\lambda$ values over the course of fine-tuning. 
 As observed, as the $\lambda$ decreases, the loss for the original dataset O (Scanner A or Institution 1) increases (Fig. \ref{fig:lossplot_ewc}a, \ref{fig:lossplot_ewc}b left plot), validating the decrease in performance on O that was observed with decreasing $\lambda$ values (red or blue curve in Fig \ref{fig:kappa_line_plot}a, \ref{fig:kappa_line_plot}b iii).
 On the other hand, for the target dataset T (Scanner B or Institution 2), we see a smooth decreasing loss curve with decreasing $\lambda$, hence validating the increase in performance with decreasing $\lambda$ (orange or green curve in Fig \ref{fig:kappa_line_plot}a, \ref{fig:kappa_line_plot}b iii). 

\subsection{Reverse Experiments}

\begin{figure}[H]
\centering
\includegraphics[clip, width=\columnwidth]{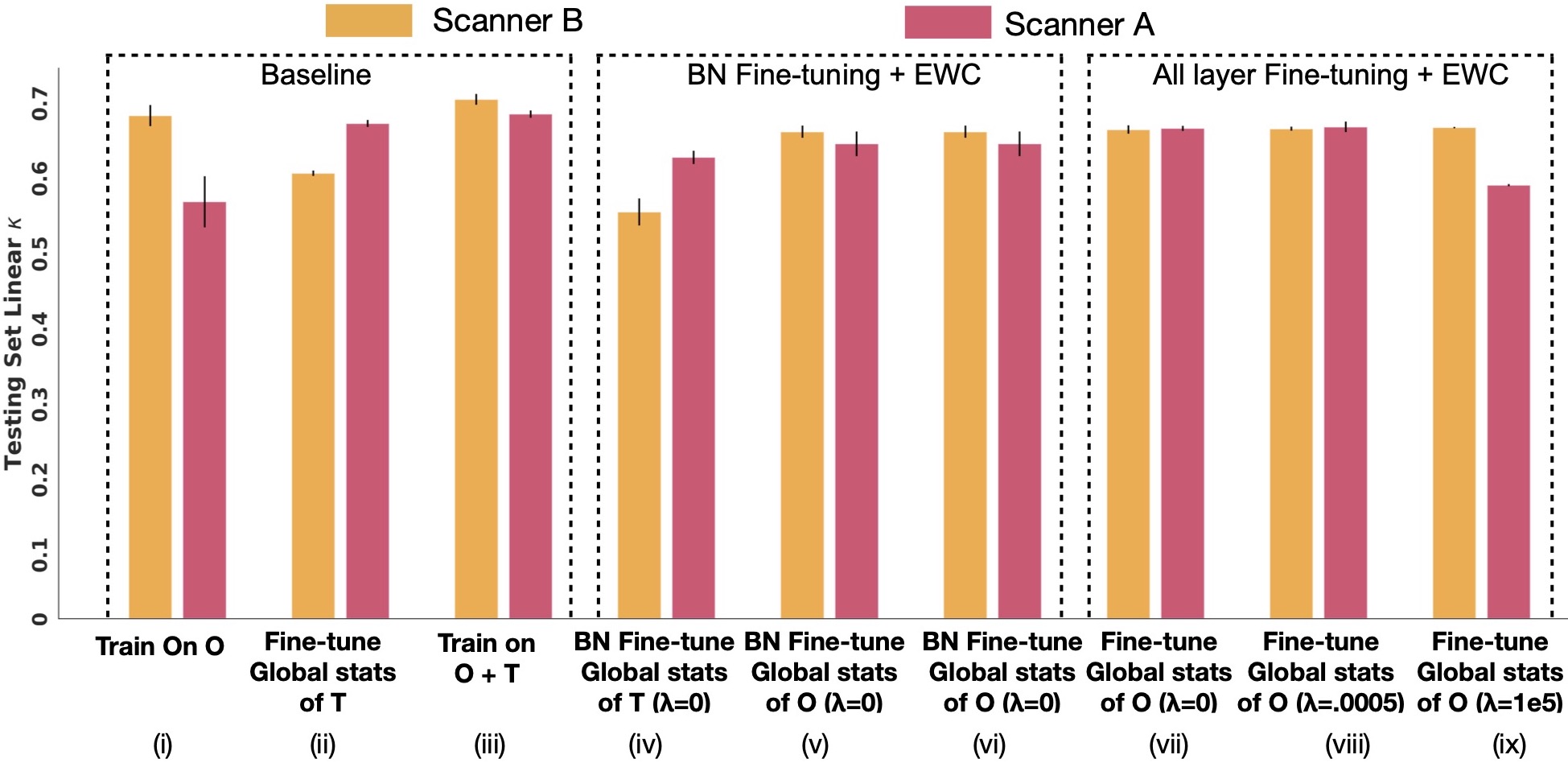}
\caption{Results for domain expansions across digital mammography systems (Scanner B to Scanner A). The different panels correspond to experiments for 1) Baseline models, 2) Only BN-layer fine-tuning with and without EWC\cite{kirkpatrick2017overcoming} and 3) All layer fine-tuning with and without EWC\cite{kirkpatrick2017overcoming}. }
\label{fig:barplot1_suppli}
\end{figure}

\begin{figure}[H]
\centering
\includegraphics[clip, width=\columnwidth]{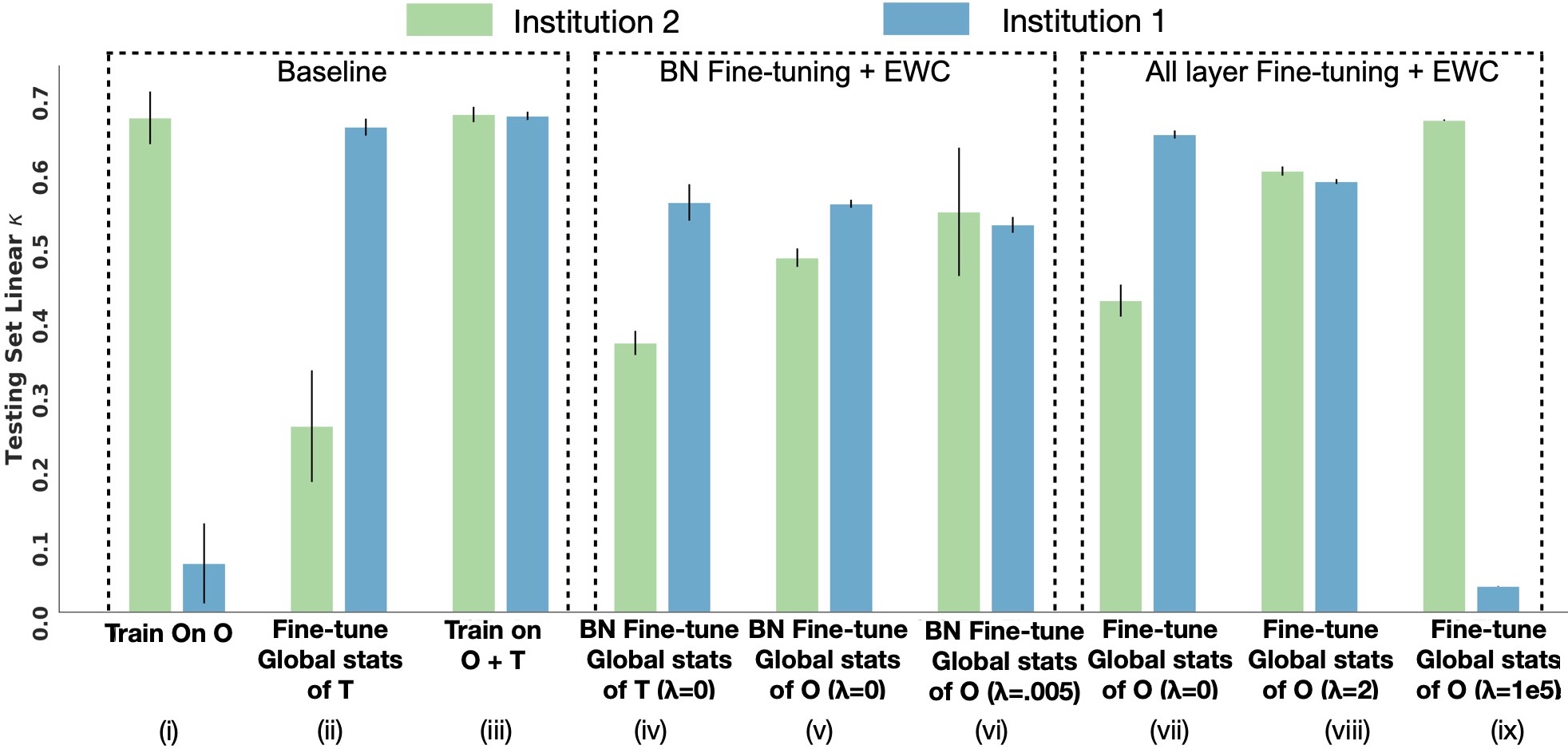}
\caption{Results for domain expansions across institutions ( Institution 2 to  Institution 1). The different panels correspond to experiments for 1) Baseline models, 2) Only BN-layer fine-tuning with and without EWC\cite{kirkpatrick2017overcoming} and 3) All layer Fine-tuning with and without EWC\cite{kirkpatrick2017overcoming}.}
\label{fig:barplot2_suppli}
\end{figure}

\begin{figure}[H]
\centering
\includegraphics[clip, width=\columnwidth]{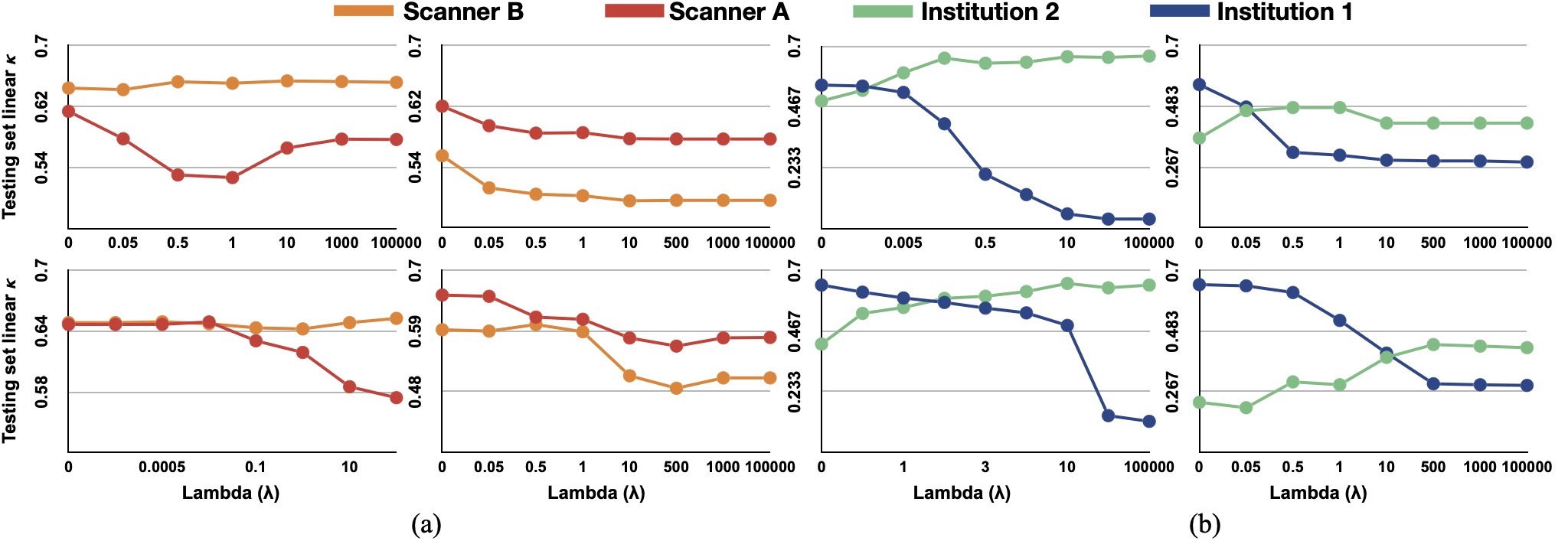}
\caption{Model performance while varying $\lambda$ when fine-tuning only BN layers (top) vs.\ all layers (bottom) with EWC\cite{kirkpatrick2017overcoming} on (a) Scanner B and Scanner A using global statistics of Scanner B (left) vs.\ Scanner A (right) ; (b)  Institution 2 and  Institution 1 using global statistics of  Institution 2 (left) vs.\  Institution 1 (right). Here, Scanner B and  Institution 2 are used as the original dataset O while Scanner A and Institution 1 are used as the target dataset T. }\label{fig:kappa_line_plot_suppli}
\end{figure}

In order to further substantiate our hypothesis, we additionally repeat all previous experiments in reverse order, \ie we consider Scanner B or Institution 2 as the original dataset O and Scanner A or Institution 1 as the target dataset T. Visualizing figures (Fig. \ref{fig:barplot1_suppli}, \ref{fig:barplot2_suppli} and \ref{fig:kappa_line_plot_suppli}) we observe similar conclusions for the reverse experiments as observed in the original order.

As for domain expansion from Scanner B (O) to Scanner A (T) is concerned, using global BN statistics of O instead of T, improves performance for both scanner types under the two scenarios: BN fine-tuning (Fig. \ref{fig:barplot1_suppli} iv, v) and all layer fine-tuning (Fig. \ref{fig:barplot1_suppli} ii,vii). Remarkable, high performance is maintained for Scanner B (O) even without the usage of EWC both for BN Fine-tuning (Fig. \ref{fig:barplot1_suppli} v, vi) as well as all layer fine-tuning (Fig. \ref{fig:barplot1_suppli} vii, viii). This can also be seen in Fig. \ref{fig:kappa_line_plot_suppli}a left column, where performance stays consistent for Scanner B (O) with increase in $\lambda$ parameter even though performance of Scanner A (T) drops. A possible reason could be that Scanner A (T), with a large dataset size ($n=59411$ for Scanner A vs.\ $n=32928$ for Scanner B) and some overlap in intensity histogram with Scanner B (O) as shown in Fig. \ref{fig:intensity_histogram}a, potentially helps in fine-tuning a model that is robust enough to scans from original scanner type, Scanner B (O) even when $\lambda=0$. However, note that this is only applicable when we use global BN statistics of O. With global BN statistics of T, we observe a drastic dip in Scanner B's (O) performance even for high $\lambda$ values (Fig. \ref{fig:kappa_line_plot_suppli}a right column).

On the other hand, in case of domain expanding from Institution 2 (O) to Institution 1 (T) (Fig. \ref{fig:barplot2_suppli} ii), catastrophic forgetting is lower than when domain expanding from Institution 1 (O) to Institution 2 (T) (Fig. \ref{fig:barplot2} ii). Furthermore, at very high values of $\lambda$ ($\lambda$=1e+5), performance on Institution 1 (T) becomes significantly lower (Fig. \ref{fig:barplot2_suppli} ix) as compared to the performance on target domain when domain expanding from Institution 1 (O) to Institution 2 (T) (Fig. \ref{fig:barplot2} ix). Similar to domain expansion from Scanner B to Scanner A, a possible explanation for this finding may be that Institution 2 (O) is smaller and less heterogeneous than Institution 1 (T). Hence while in the first case (Fig. \ref{fig:barplot2_suppli} ii), when the model is fine tuning on Institution 1 (T), it possibly learns a more robust model due to dataset size and heterogeneity, that improves the model's ability to generalize well on Institution 2 (O). In the second case (Fig. \ref{fig:barplot2_suppli} ix), having too high a $\lambda$ essentially allows it to learn only from the Institution 2 dataset, which is smaller and more homogeneous, resulting in poor performance on Institution 1. However, comparing Fig. \ref{fig:kappa_line_plot_suppli}b left and right column, we can clearly ascertain that global BN statistics of O together with EWC is necessary for the model to avoid catastrophic forgetting on Institution 2 (O). Additionally, fine-tuning all layers is needed to achieve high performance on Institution 1 (T).

\subsection{State Of The Art Comparisons}\label{subsec:SoTA}
In the results section, We have compared the results of our approach with two existing state-of-the-art techniques: BN fine-tuning \cite{karani2018lifelong} and baseline EWC \cite{kirkpatrick2017overcoming}. For additional comparison, we trained models on Institution 1 and Institution 2 with Zenke et. al.’s code\footnote{\url{https://github.com/ganguli-lab/pathint}} \cite{zenke2017continual}, resulting in a $\kappa$= -0.054 on Institution 1 and $\kappa$=0.573 on Institution 2 ($p < 0.01$), which shows a lack of effectiveness in mitigating catastrophic forgetting for our real world dataset. Despite rigorous experimentation with the \textit{strength} parameter (denoted by $c$ in the paper \cite{zenke2017continual}), which controls the trade-off between memory retention and learning rate, we failed to observe any improvement in performance on Institution 1. 

We did not compare our results with more recent continuous learning techniques, such as OWM \cite{zeng2019continual}, which are limited to extracting features from a model trained on the “combined dataset”. This was out of scope for our problem because data between different medical institutions is not allowed to be shared due to patient privacy regulations. 

\begin{figure}[H]
\centering
\includegraphics[clip, width=\columnwidth]{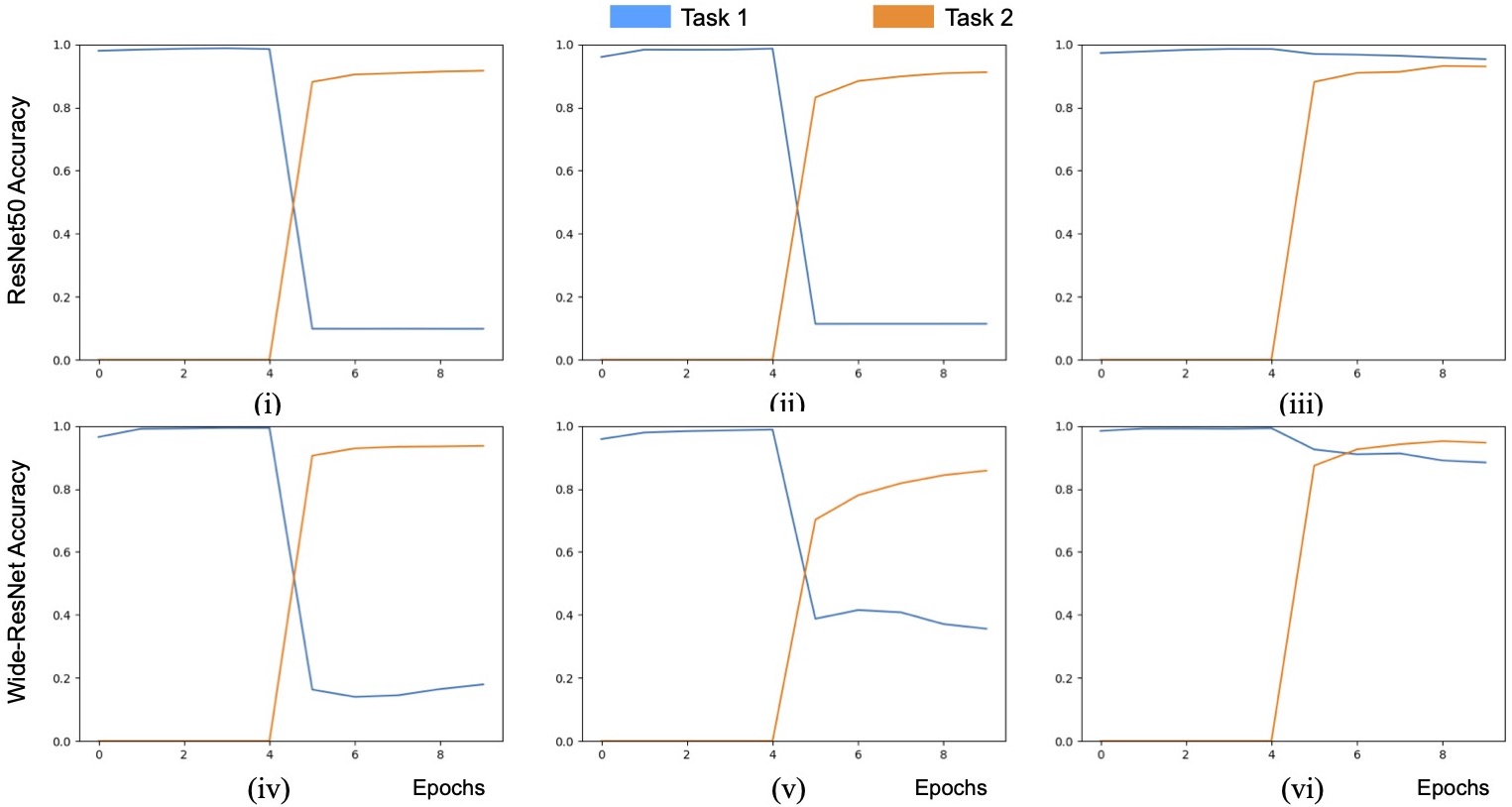}
\caption{Accuracy of a ResNet50 architecture(top) (i) standard training (ii) using dropout probability of 0.5 before the last layer (iii) using batch statistics of task 1 while fine-tuning; a Wide-ResNet architecture(bottom) (iv) standard training (v) using dropout probability of 0.1 (vi) using batch statistics of task 1 while fine-tuning }
\label{fig:sota}
\end{figure}

We did, however, compare our results with recent work by \cite{mirzadeh2020dropout}, in which they show how dropout implicitly produces a gating mechanism in the network that can address catastrophic forgetting. Since their results were only demonstrated for a multi-layer perceptron (MLP), we attempted to adapt their idea for more modern, deeper network architectures, such as ResNet50, in order to accommodate larger and more heterogeneous datasets. We trained a ResNet50 model on two tasks of permuted MNIST described in their paper. The permuted MNIST dataset is generated by shuffling pixels such that the permutation is the same between images of the same task but is different across the tasks \cite{goodfellow2013empirical}. When a standard Resnet50 architecture is trained on task 1 and fine-tuned on task 2, performance on task 1 drops sharply (as shown in Fig. \ref{fig:sota} i). Although we were able to replicate their results (perform domain expansion with minimal catastrophic forgetting) using a two layered MLP (with dropout), we were unable to achieve high performance using a Resnet50 architecture. Indeed, \cite{farquhar2019robust} previously criticized Permuted MNIST for being an “unrealistic best case scenario for continuous learning”, since in a real world, new dataset would have some concurrence with the old dataset. In order to adapt \cite{mirzadeh2020dropout} to our Resnet50 architecture, we inserted dropout with probability of 0.5 before the last layer. This led to significant levels of catastrophic forgetting with accuracy of 0.12 and 0.91 on validation datasets from task 1 and 2 respectively after training on the second task as shown in Fig. \ref{fig:sota} ii. This shows that while their dropout technique works well with a shallow network like MLP, the technique fails for complex model architectures \footnote{One of the possible reasons can be that MLP, due to its fully connected layers, is somewhat blind to the permutations and hence doesn't forget much from task 1 when trained on task 2.}. However, when a ResNet50 architecture was fine-tuned on task 2 using global BN statistics of task 1, catastrophic forgetting was significantly mitigated. Performance on task 1 and task 2 was 0.95 and 0.93 respectively as shown in Fig. \ref{fig:sota} iii.

Moreover, in order to test dropouts in the intermediate convolution layers, we additionally experimented with Wide ResNet architecture from \cite{zagoruyko2016wide}, given that they could successfully incorporate dropouts inside each residual blocks (after every convolutional layer). We used the same tasks 1 and 2 from Permuted MNIST as used for Resnet50 architecture. We first trained a vanilla Wide ResNet architecture (without dropout) on task 1 and subsequently fine-tuning on task 2 that led to drastic decrease in performance on task 1 as shown in Fig. \ref{fig:sota} iv. The model attained accuracies of 0.18 and 0.94 on task 1 and task 2, respectively. Adding dropouts to the wide residual network, similar to \cite{zagoruyko2016wide} with widening factor of 2 and 50 convolution layers (WRN-50-2), resulted in considerable levels of catastrophic forgetting as shown in Fig. \ref{fig:sota} v. For a dropout probability of 0.10, accuracies for task 1 and 2 were 0.36 and 0.86 respectively. At higher dropout probabilities, the model was unable to converge for task 2. As shown in Fig. \ref{fig:sota} vi, when the model was fine-tuned using global BN statistics of task 1, performance on task 1 and task 2 was 0.88 and 0.95, respectively, hence alleviating catastrophic forgetting.
In conclusion, our results show that all the above state-of-the-art techniques are unable to alleviate catastrophic forgetting for modern CNN architectures on commonly used Permuted MNIST dataset even with extensive hyper-parameter tuning. However using the global BN statistics technique, we can easily mitigate this catastrophic forgetting to a large extent. 

